\documentclass[a4paper, 10pt, conference]{ieeeconf}      
\usepackage{FG2024}

\FGfinalcopy 

\usepackage{epsfig}
\usepackage{graphicx}
\usepackage{amsmath}
\usepackage{amssymb}
\usepackage{textcomp}
\usepackage{xcolor}
\usepackage{longtable}
\usepackage{subcaption}
\usepackage{dblfloatfix}
\usepackage{float}
 \usepackage{booktabs}
\usepackage{bigstrut}
\usepackage{multirow}
\usepackage{caption}
\usepackage{fancyhdr}
\usepackage{listings} 
\usepackage{changepage}
\usepackage{pgfplots}
\usepackage{balance}
\usepackage{pifont}
\PassOptionsToPackage{hyphens}{url}\usepackage{hyperref}
\def\FGPaperID{0073} 

\title{\LARGE \bf
VoxAtnNet: A 3D Point Clouds Convolutional Neural Network for Generalizable Face Presentation Attack Detection}

\author{\parbox{16cm}{\centering
    {\large Raghavendra Ramachandra$^1$ \ Narayan Vetrekar$^2$ \ Sushma Venkatesh$^3$ \ Savita Nageshker$^2$ \  Jag Mohan Singh$^1$ \ R. S. Gad$^2$}\\
    {\normalsize
    $^1$ Norwegian University of Science and Technology (NTNU), Gj{\o}vik, Norway\\
    $^2$ School of Physical and Applied Sciences, Goa University, Goa, India\\
    $^3$ AiBA AS, Gj{\o}vik, Norway}}
}

\begin{document}

\ifFGfinal
\thispagestyle{empty}
\pagestyle{empty}
\else
\author{Anonymous FG2024 submission\\ Paper ID \FGPaperID \\}
\pagestyle{plain}
\fi
\maketitle

%
%
%
%
%
%
\begin{abstract}

Facial biometrics are an essential components of smartphones to ensure reliable and trustworthy authentication. However, face biometric systems are vulnerable to Presentation Attacks (PAs), and the availability of more sophisticated presentation attack instruments such as 3D silicone face masks will allow attackers to deceive face recognition systems easily. In this work, we propose a novel Presentation Attack Detection (PAD) algorithm based on 3D point clouds captured using the frontal camera of a smartphone to detect presentation attacks. The proposed PAD algorithm, \textit{VoxAtnNet}, processes 3D point clouds to obtain voxelization to preserve the spatial structure. Then, the voxelized 3D samples were trained using the novel convolutional attention network to detect PAs on the smartphone. Extensive experiments were carried out on the newly constructed 3D face point cloud dataset comprising  bona fide and two different 3D PAIs (3D silicone face mask and wrap photo mask),  resulting in 3480 samples. The performance of the proposed method was compared with existing methods to benchmark the detection performance using three different evaluation protocols. The experimental results demonstrate the improved performance of the proposed method in detecting both known and unknown face presentation attacks. 

\end{abstract}

\section{INTRODUCTION}
Face biometrics are the primary building blocks in smartphone-based applications that require secure and trustworthy authentication. Smartphone applications include the unlocking of phones, downloading applications, banking transactions, and finance applications.  The wide adoption of face biometrics can be attributed to the highly accurate performance and usability that are essential in smartphone applications. The popularity of face biometrics has resulted in the deployment of more than 96 million smartphones as of 2019, and is expected to grow to 800 million smartphones by 2024 \cite{SmarptMar}.

The wide deployment of face recognition systems on smartphones has enabled several applications, but at the same time suffered a series of presentation (or spoofing) attacks.  The goal of Presentation Attacks (PAs) is to present the facial artefact (or Presentation Attack Instrument (PAI)) of a legitimate user to gain unauthorized access to smartphones or smartphone applications.  The PAI can be generated using different types of artifact materials such as print attack, electronic display attack, wrap attack (in which printed photo is wrapped around the face), and 3D silicone face mask attack. A recent analysis presented by the consumer champion \cite{SmrtSpoof} which tested 43 new smartphones released in 22, shows that 40\% (19 smartphones) can be easily spoofed using a simple print attack. Therefore, the detection of PAs is of paramount importance to enable trustworthy authentication in smartphone applications.

\begin{figure*}[t!]
\begin{center}
\includegraphics[width=1.0\linewidth]{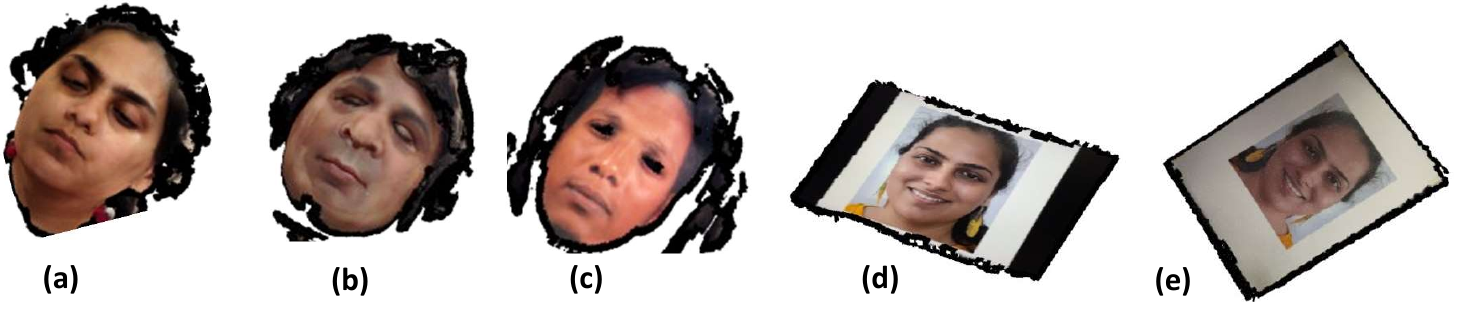}
\end{center}
   \caption{Example of 3D point clouds captured using frontal camera of Apple iPhone 12 Pro (a) Bona fide (b) 3D silicone face mask (c) wrap paper attack (d) print paper attack (e) display attack.  It can be noted that, the use of 2D artefacts like   print paper attack and  display attack are easy to detect due to the lack of depth information. }
\label{fig:intro}
\end{figure*}

Face Presentation Attack Detection (PAD) are extensively studied in the literature that has resulted in several techniques \cite{ramachandra2017presentation, yu2022deep} on the smartphone data. The fundamental challenge of achieving the reliable face PAD is to develop the system that is agnostic to the different materials that are used to generate the PAs. Therefore, detecting the un-known attack instrument is an essential contribution for the success of the PAD algorithms. Furthermore, the developed PAD algorithms must be robust to the environmental changes and skin tone to achieve the generalizability across different capture conditions and demographics.

Existing face PAD techniques can be broadly classified as: texture based, frequency based, motion or temporal based and deep learning-based approaches. Early PAD approaches were based on the use of Local Binary Patterns (LBP) \cite{maatta2011face, boulkenafet2016face1}, Histogram of Oriented Gradients (HOG)  \cite{HOGPAD}, Speeded Up Robust Features (SURF)  \cite{boulkenafet2016face}, Binarised Statistical Image Features (BSIF) \cite{BSIFFacePAD} and Scale-Invariant Feature Transform (SIFT)  \cite{patel2016secure}. The extracted texture features are classified using a binary classifier, such as an Support Vector Machine (SVM), to make the final decision. Although texture-based features have indicated acceptable performance on known attacks, they fail to detect an unseen attacks.  The limitations of texture-based approaches have led to the other types of features based on frequency or time-frequency features \cite{BSIFFacePAD, Moier2015face}. Although these features have shown robustness in detecting electronic display attacks, their detection performance drops with the other types of attacks. The use of temporal or motion features have been extensively studied to detect PAs, especially use of remote photoplethysmography (rPPG) \cite{yu2021transrppg} has indicated promising results for detecting print attacks. However, detection accuracy decreases with video replay attacks, and these methods are computationally expensive and sensitive to variations in external lighting.

Deep learning-based PAD techniques have been widely investigated. Earlier deep learning-based approaches were based on using pre-trained networks \cite{yang2014learn} for feature extraction, which were then classified to detect PAs.  The fusion of multiple pre-trained CNNs was presented in \cite{ramachandra2020face}, which indicated superior performance on a smartphone dataset collected with different phones and environmental conditions. Later deep-learning-based techniques are based on multichannel CNN \cite{george2019biometric}, pixel-based supervision \cite{george2019deep}, attention model \cite{wang2019multi, sun2022danet}, and transformer models \cite{TransformerPAD} which have indicated improved performance with known attacks, while these techniques still struggle to indicate the robustness to unknown attacks. Recently, face PAD methods has introduced, advanced deep learning techniques that have been developed based on the domain adaptation \cite{zhang2021face, liu2022contrastive, wang2020cross, yu2021salience, shao2019multi, zhang2022effective}, self-supervised learning \cite{SelfSuperPADRaghu} and meta-learning \cite{shao2020regularized}.  However, these PAD methods are computationally expensive and require more data to achieve a good detection performance for unseen attacks.

The use of auxiliary information, such as depth, is well explored for face PAD from smartphone data. Early work employing depth information was presented in \cite{atoum2017face} where holistic depth maps and patch-based images were learned using two-stream CNNs. Binary supervision based on the depth and rPPG signal was presented using the CNN-RNN architecture \cite{liu2018learning}, which indicated a marginal improvement in the cross-dataset evaluation.   The depth-supervised residual spatial gradient block was introduced in \cite{wang2020deep} using contrastive depth loss, which indicated improved performance, particularly in photo attacks. The generation of 3D point clouds from a given video was proposed in \cite{li20203dpc}, which indicated improved performance on 2D based attack instruments. Although the reconstruction of depth used as auxiliary information indicates a marginal improvement in detection performance, the reconstructed depth represents pseudo-depth information. It is worth noting that pseudo 3D information does not reflect the real depth information, and is tricked by the attacker by introducing pseudo depth while presenting the attack.

Recently, the first work on using point clouds for face PAD was presented in \cite{PointcloudnetPAD} in which the iPad was used to collect LiDAR data (from the back camera of the iPad) of both bona fide and PA. Three different types of PAs were used, namely, print paper, display screen, and rigid face mask, which were collected under different lighting conditions.  A two-stream serial CNN architecture was proposed to combine RGB and (point cloud \& depth) data to detect presentation attacks. The obtained results on 12 data subjects indicate that point clouds are less variant to lighting conditions and improves the detection accuracy when trained and tested with the same type of attacks.   Thus, the results reported in \cite{PointcloudnetPAD} did not report the detection performance for unseen attacks, and PAs were limited to 3D rigid masks. Furthermore, the detection performance did not show significant differences between RGB and point clouds, which can be attributed to low-quality PA instruments. 

\begin{figure*}[t!]
\begin{center}
\includegraphics[width=1.0\linewidth]{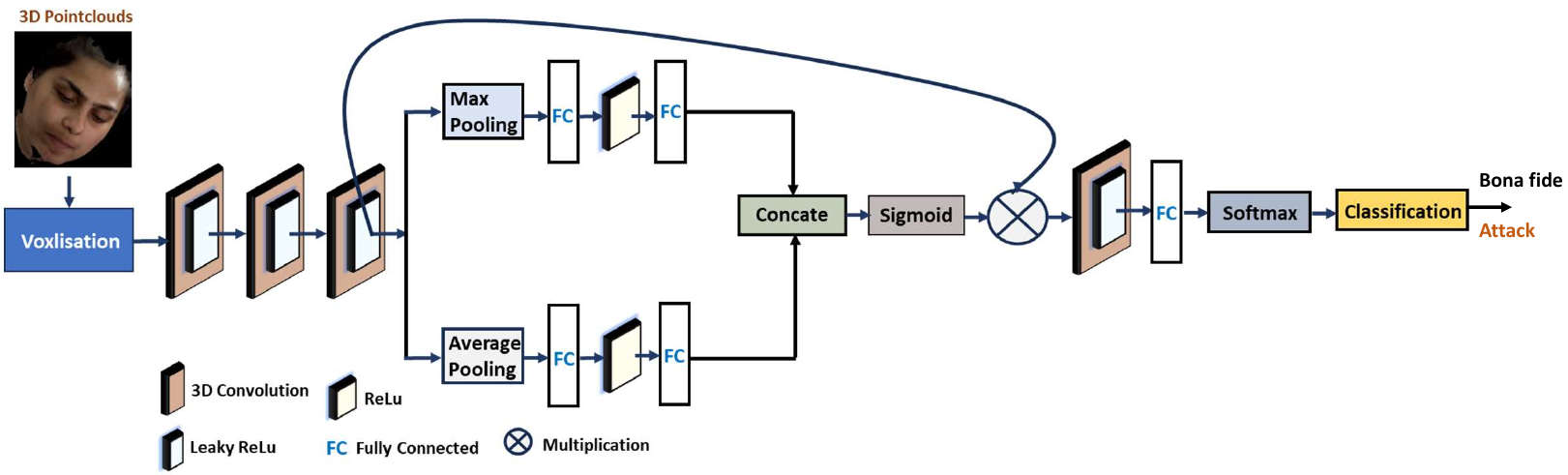}
\end{center}
   \caption{Block diagram of the proposed  VoxAtnNet for face PAD. The novelty of the VoxAtnNet includes the voxelization and the attention module with skip connections. }
\label{fig:Prop}
\end{figure*}

In this work, we present a first work on detecting the face PAs on the smartphones using point cloud data processing that are collected using the frontal camera of Apple iPhone 12Pro (refer Figure \ref{fig:intro} for example 3D Point clouds captured). We consider 3D PAIs such as silicone face mask and wrap paper as other type of PAIs like print and electronic display attack are 2D based and does not indicate the depth as shown in the Figure \ref{fig:intro}. Therefore, 2D based PAIs are easy (and obvious) to detect using point clouds.   The point clouds were processed by performing voxelization to ensure a rich spatial structure, which was then fed to the proposed 3DCNN network to detect the PAs. Compared to \cite{PointcloudnetPAD}, the proposed method is different: (1) The point clouds are captured using a frontal camera; thus, the user can self-capture the face samples and thus improve usability. (2)  The proposed method employs voxelization and 3DCNN instead of a multistream network that considers RGB and depth separately. Therefore, the proposed method is based on a single stream CNN that can directly process the point clouds. (3) Benchmarking results for unseen PA attack scenarios with 3D PAIs. The main contributions of this study are as follows: 

\begin{itemize}
\item Proposed a novel approach to detect face presentation attacks using 3D point clouds recorded from the frontal camera of a smartphone. To the best of our knowledge, this is the first study to explore the utility of 3D point clouds (dense) from the frontal camera of a smartphone for facial PAD. 
\item Proposed a new algorithm based on voxelization and 3DCNN attention model for reliable detection of the unseen presentation attack instruments.
\item New point cloud dataset collected using Apple iPhone 12 Pro with 30 bona fide subjects resulting in 1014 point clouds and two different PAI such as wrap photo attack and 3D Silicone mask. The entire dataset comprised of 3480 3D point cloud samples. This database will be made available for research purposes (\url{https://sites.google.com/view/speciblab/research/3d-pcpa?authuser=1}). 
\item Extensive experiments are  performed to benchmark the detection performance of the proposed method with the four different existing point cloud based methods.  
\end{itemize}

The rest of the paper is organized as follows: Section \ref{sec:Pro} presents the proposed 3D point cloud based face PAD technique, Section \ref{sec:DB} presents the newly constructed 3D point cloud presentation attack datasets using a 3D silicone face mask and wrap photo attack, Section \ref{sec:Exp} presents the quantitative results of the proposed method and point-cloud-based PAD techniques on different experimental evaluation protocols, Section \ref{sec:limit} discuss the limitations of the present work, and Section \ref{sec:Cocn} concludes the paper.   

\section{Proposed VoxAtnNet 3D Face PAD}
\label{sec:Pro}

Designing a network involves infinite choices of parameters; however, earlier works \cite{maturana20153d}, \cite{maturana2015voxnet} have explored different 3D CNN architectures on point clouds, indicating that the use of low-parameterized models can lead to the best performance. Therefore, we are motivated to propose a novel lightweight Convolutional Neural Network (CNN) for point cloud based face PAD. Figure \ref{fig:Prop} shows a block diagram of the proposed method (VoxAtnNet) for 3D face-presentation attack detection. The proposed VoxAtnNet  process the 3D point clouds as a volumetric occupancy grid (voxels) using 3D convolutional attention networks to reliably detect the face presentation attack detection. VoxAtnNet can be structured using two building blocks: (a) voxelization of 3D point clouds, and (b) a 3D convolutional attention network to detect presentation attacks.

\begin{figure*}[htp]
\begin{center}
\includegraphics[width=1.0\linewidth]{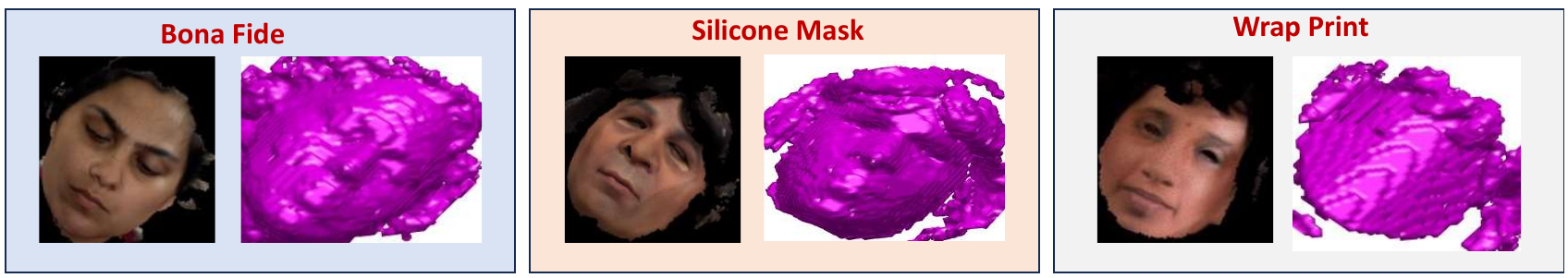}
\end{center}
   \caption{Qualitative results of the voxelization of bona fides and PAIs. The voxelization of the bona fide indicated a rich spatial structure (or high quality surface details) compared to both PAIs. The spatial structure of 3D wrap print attacks indicates a poor spatial structure  (or poor surface details) compared with a 3D silicone face mask.}
\label{fig:Vox}
\end{figure*}

Given the 3D point clouds scanned using the frontal camera of iPhone 12Pro, the first step is to represent point clouds as occupancy grids using voxelization \cite{maturana2015voxnet}. This step is important for transforming the irregular spatial sampling of a point cloud into a regular sampled structure. In this study, we employed voxelization, which allows the dense representation of point clouds and therefore results in an efficient representation of space and range measurements. Furthermore, the use of dense representation enables the utility of convolution operations as they represent the data that are sampled regularly in the spatial domain. Importantly, voxelization can capture the rich spatial structure of the face image, which can reveal the differences between real and artifact facial presentations. 
Voxelization maps each point cloud point (x,y,z) to discrete voxel coordinates (i,j,k) using uniform discretization of the voxel grid space. The voxelization outcome depends on the origin, orientation, and resolution of the voxel grid space.  In this study, we adapted the process mentioned in \cite{maturana2015voxnet} to compute the parameters. The origin was assumed as an input, and the orientation was assumed to align the grid frame with the direction of gravity. To maintain a consistent orientation of objects around the z-axis, we argue the dataset during the training by creating copies of the input point cloud by rotating it by $360 ^{\circ}$.   For resolution, we used a fixed occupancy grid of $64 \times 64 \times 64$ voxels by making sure to capture the shape information and avoid aliasing when voxels are large. Figure \ref{fig:Vox} shows an example of voxelization corresponding to the bona fide and attack samples, which shows the visual distinction in the spatial structures.

The spatial structure obtained after voxelization for the corresponding point clouds was then fed to the novel CNN architecture designed to detect face presentation attacks.  Figure \ref{fig:Prop} shows the architecture of the proposed VoxAtnNet with 23 layers connected in a residual manner. The initial or early layers of VoxAtnNet have a series connection of 3D convolution with leaky ReLU layers that can effectively encode the planes and corners at different orientations. The stacked convolution layers have different filter sizes, numbers of filters, and strides that enable the construction of hierarchical features for the input occupancy grid. The first convolution layer has a filter size of $5 \times 5 \times 5$, number of filters of 64, and stride of [2,2,2], while the second and third convolution layers have a filter size of $3 \times 3 \times 3$, number of filters of 32, and stride of [1,1,1].  In the second stage, we introduce an attention network to guide the learning to focus on the important spatial structure in the given occupancy grid. The proposed attention unit is slightly different from the existing spatial attention as the proposed method adds a series of fully connected and ReLU independently on max and average pooling, which are then concatenated. Thus, the use of fully connected and ReLu layers emphasizes the area of attention that can help VoxAtnNet to find the spoof patterns in the spatial structure of the occupancy maps. The concatenated features are then passed through the sigmoid layer and multiplied by the skip connection from the convolution-3 layer, which can further represent the rich features. The last part of the network processes the multiplied features through the serial connection of convolution (with $3 \times 3$, number of filters is 32, and stride of [1,1,1], respectively), leaky ReLU, and fully connected layer before performing classification using the softmax layer.

Training of VoxAtnNet is performed using Stochastic Gradient Descent with momentum (SGDM) with cross-entropy loss. The SGDM was initialized with a learning rate of 0.01. Data augmentation is dynamically performed during training by adding randomly perturbed copies (adding jitter noise, mirrored, and shifted) for each instance. The mini-batch size used in this study was 32. The proposed method VoxAtnNet has 35.7M learnable parameters.

\section{3D face point cloud presentation attack Dataset (3D-PCPA)}
\label{sec:DB}
\begin{figure*}[htp]
\begin{center}
\includegraphics[width=0.85\linewidth]{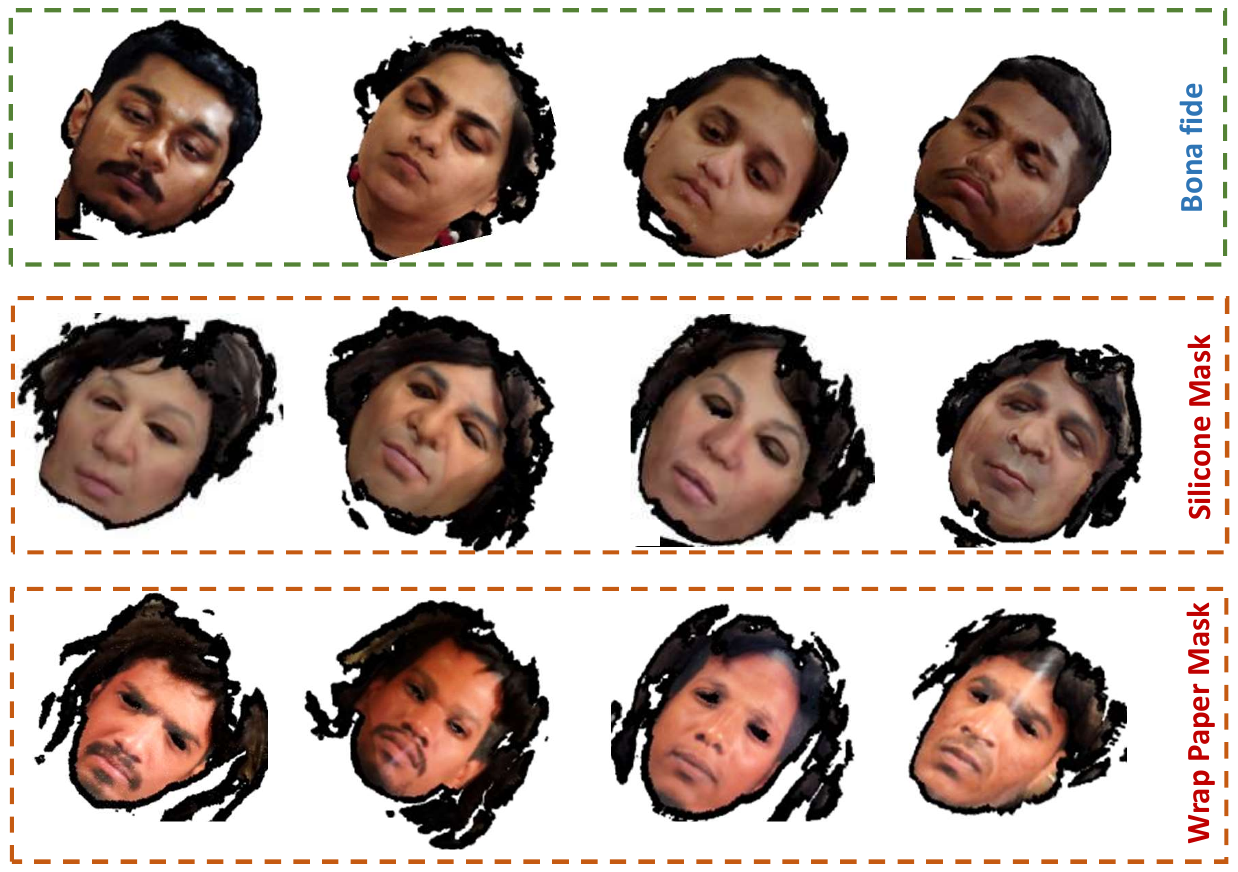}
\end{center}
   \caption{Examples 3D point clouds samples from 3D-PCPA dataset corresponding to bona fide, 3D silicone face mask and wrap paper mask. }
\label{fig:DB}
\end{figure*}

This section describes a newly constructed 3D-PCPA database acquired using an iPhone 12 Pro Smartphone. The frontal camera of the iPhone 12 Pro, in which the user interacts and captures the 3D point  clouds on their own. The 3D scan was self-captured by the user at 15-20 cms between the face and smartphone. The 3D-PCPA  database comprises bona fide and presentation attack face 3D point clouds to capture the point clouds acquired in multiple  sessions. The details related to each category of acquisition are summarized in the following subsections. The summary of number of samples acquired under each category is detailed in Table~\ref{tab:database-summary} and Figure \ref{fig:DB} shows the example point clouds from the 3D-PCPA dataset.

\begin{table}[htbp]
	\centering
	\caption{Statistics of 3D face point cloud presentation attack Dataset (3D-PCPA)}
	\resizebox{0.48\textwidth}{!}{
		\begin{tabular}{|c|c|}
			\hline
			\textbf{Data Type} & \multicolumn{1}{l|}{\textbf{Number of samples}} \bigstrut\\
			\hline
			Bona fide & 1014 \bigstrut\\
			\hline
			3D Silicone Mask Artefacts & 840 \bigstrut\\
			\hline
			3D Wrap Photo Artefacts & 1626 \bigstrut\\
			\hline\hline
			\textbf{Total} & \textbf{3480} \bigstrut\\
			\hline
		\end{tabular}%
		\label{tab:database-summary}%
}
\end{table}%

\subsection{$Bona fide$ subset of 3D-PCPA Database}
\label{ssec:bonafide}
The $bona fide$ face point clouds of the 3D-PCPA database were collected from $30$ different data subjects ($16$ male and $14$ female) in an indoor office environment. Each subject was asked to scan his/her own face using a smartphone in multiple sessions, and data collection was performed over 1-3 weeks’ time.  A total of $1014$ face point clouds were acquired, which corresponded to 30 to 33 3D point cloud samples per data subject.    
 \subsection{Presentation attack subsets for 3D-PCPA database}
\label{ssec:presentation attack}
In this work, we tested 3D PAIs over 2D PAIs to effectively address the attack potential of 3D PAD techniques for 3D PAIs.  Therefore, we considered two different 3D PAIs: (a) a 3D silicone mask and (b) a 3D wrap paper photo attack. The 3D silicone masks used in this work are custom-made high-quality face masks that have higher vulnerabilities to  Face Recognition systems (FRS). Wrap paper photo attacks are generated by wrapping the print photo on the attacker’s face to simulate the pseudo depth.  

 \textbf{3D Silicon Face Mask Artefact:} We have employed four ($2$ male and $2$ female subject) unique 3D silicone face mask that are used to capture the 3D point clouds. These 3D silicone masks were worn by the data subjects, and scans were performed on their own to capture the 3D point clouds. Data collection was carried out in different sessions for a duration of three weeks, where four masks were worn by 15 different data subjects. In total, the 3D silicon PAI had 840 point cloud scans corresponding to four unique masks.  

\textbf{3D Wrap photo Artefacts:} The 3D wrap photo print attacks generation has three steps (1) We capture the high resolution photo of 15 different data subjects ($10$ male and $5$ female) using DSLR camera (2) Digital photo is printed on a high quality papers using color laser printer (Model:Konica Minolta's bizhub C360i). (3) The printed photo was then wrapped around the face and self-captured using the frontal camera of the smartphone to obtain point clouds. The attacks were generated by wearing these 15 unique wraps by 20 different data subjects in multiple sessions varying from 1 to 3 weeks, resulting in  1626 3D wrap photo artefacts.

\section{Experiments and Results}
\label{sec:Exp}
In this section, we present the performance evaluation protocol and quantitative results of the proposed 3D PAD method on the newly collected 3D-PCPA dataset.  Because the 3D point clouds were employed directly to detect the face PAD, we compared the detection performance of the proposed method with three different well established point clouds based classification techniques such as PointNet \cite{qi2017pointnet}, PointNet++  \cite{qi2017pointnet++}, VoxNet \cite{maturana2015voxnet} and Masked AutoEncoders (Point-MAE) \cite{pang2022MAE}.  The quantitative performance of the  3D PAD algorithms was evaluated using ISO/IEC  30107- 3 \cite{ISO-IEC-30107-3-PAD-metrics-170227} metrics. The `Attack Presentation Classification Error Rate (APCER) is defined as the proportion of attack presentations incorrectly classified as bona fide, and BPCER is defined as the portion of the bona fide incorrectly classified as attack presentation' \cite{ISO-IEC-30107-3-PAD-metrics-170227}. In addition, we present quantitative results using the Detection-Equal Error Rate (D-EER) and Detection Error Trade-off (DET) curves.

\subsection{Performance evaluation protocol}
The performance of the 3D PAD techniques was evaluated by dividing the 3D-PCPA datasets into two disjoint sets, namely, the training and testing sets.  Bona fide data with 30 unique data subjects were portioned to have 20 unique data subjects in the training set and the remaining 10 in the testing set. This resulted in 655 point cloud samples in training and 359 point cloud samples during testing. For the 3D silicone masks, we chose two unique masks (one male and one female) for training and the remaining two unique masks for the testing set. This resulted in 638 point cloud samples in training and 202 point cloud samples during testing. For 3D wrap photo masks, out of 15 unique wrap photo artefacts, eight (5 male and 3 female) were selected for training, and the remaining seven (5 male and 2 female) were selected for the testing set. This resulted in 769 point cloud samples in the training set and 857 point cloud samples in the testing set.  Table \ref{tab:train-test} shows the statistics of the train and test partitions. 
\begin{table}[htp]
\centering
\caption{Training and testing partition of 3D-PCPA dataset}
\label{tab:train-test}
\resizebox{\columnwidth}{!}{%
\begin{tabular}{|c|c|c|}
\hline
\textbf{Data Type} &
  \textbf{\begin{tabular}[c]{@{}c@{}}Number of  \\  training samples\end{tabular}} &
  \textbf{\begin{tabular}[c]{@{}c@{}}Number of \\ testing   samples\end{tabular}} \\ \hline
Bona fide                                                                & 655 & 359 \\ \hline
\begin{tabular}[c]{@{}c@{}}3D silicone  \\  masks Artefacts\end{tabular} & 638 & 202 \\ \hline
\begin{tabular}[c]{@{}c@{}}3D wrap \\ photo   Artefacts\end{tabular}     & 769 & 857 \\ \hline
\end{tabular}%
}
\end{table}

\begin{figure*}[htp] 
    \centering
    \subfloat[PointNet \cite{qi2017pointnet} ]{%
        \includegraphics[width=0.2\textwidth]{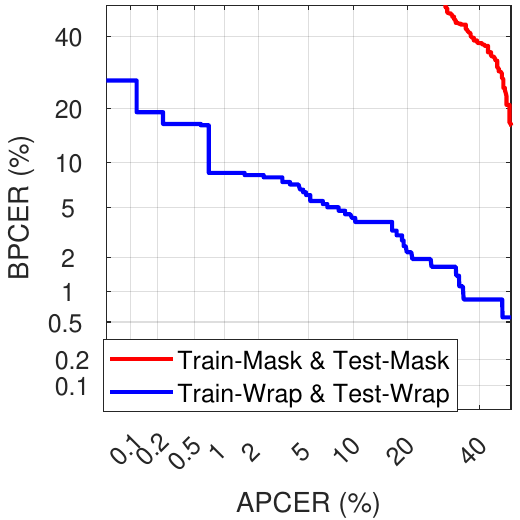}%
        \label{fig:Intra_a}%
        }%
        \hfill%
    \subfloat[PointNet++ \cite{qi2017pointnet++}]{%
        \includegraphics[width=0.2\textwidth]{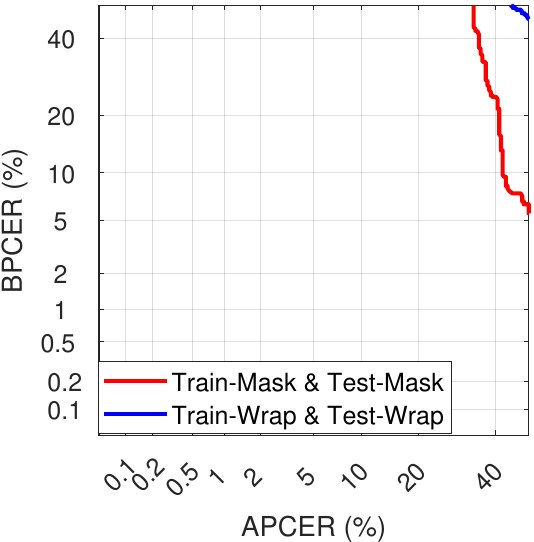}%
        \label{fig:Intra_b}%
        }%
        \hfill%
    \subfloat[Point-MAE \cite{pang2022MAE}]{%
        \includegraphics[width=0.2\textwidth]{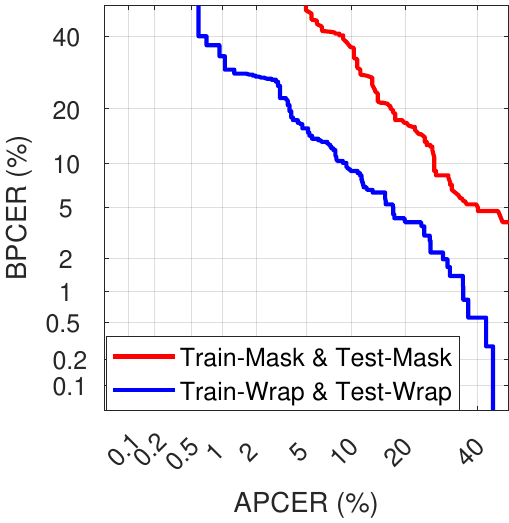}%
        \label{fig:Intra_c}%
        }%
         \hfill%
    \subfloat[VoxNet \cite{maturana2015voxnet}]{%
        \includegraphics[width=0.2\textwidth]{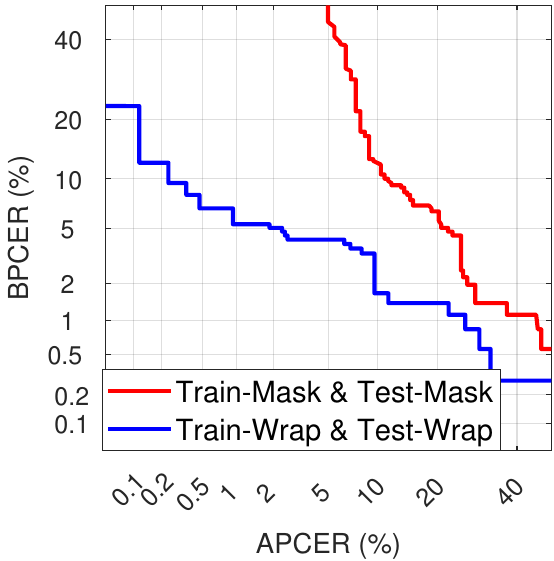}%
        \label{fig:Intra_d}%
        }%
    \hfill%
    \subfloat[Proposed Method]{%
        \includegraphics[width=0.2\textwidth]{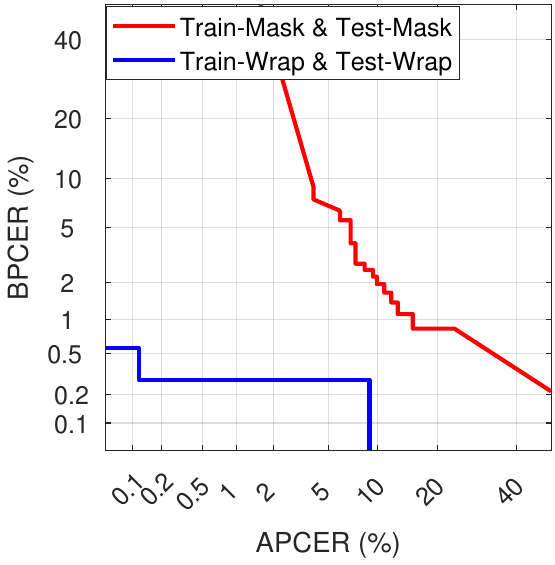}%
        \label{fig:Intra_e}%
        }%
    \caption{DET curves showing the detection performance with Intra protocol (best viewed in color). X-Axis indicates the APCER and y-axis indicates the BPCER. 
    }
\end{figure*}

\begin{table*}[]
\centering
\caption{Quantitative results of the proposed VoxAtnNet and existing methods on different evaluation protocol. The proposed VoxAtnNet indicates the best detection performance compared to existing methods.}
\label{tab:Resu}
\resizebox{1.9\columnwidth}{!}{%
\begin{tabular}{|c|c|c|c|c|cc|}
\hline
\multirow{2}{*}{\textbf{Evaluation   Protocol}} &
  \multirow{2}{*}{\textbf{Training set}} &
  \multirow{2}{*}{\textbf{Testing set}} &
  \multirow{2}{*}{\textbf{3D PAD   Algorithms}} &
  \multirow{2}{*}{\textbf{D-EER (\%)}} &
  \multicolumn{2}{c|}{\textbf{BPCER @   APCER =}} \\ \cline{6-7} 
                       &                       &                       &                            &               & \multicolumn{1}{c|}{\textbf{10\%}} & \textbf{5\%}   \\ \hline
\multirow{8}{*}{Intra} & \multirow{4}{*}{Mask} & \multirow{4}{*}{Mask} & Point-MAE \cite{pang2022MAE}                 & 17.84         & \multicolumn{1}{c|}{37.14}         & 58.77          \\ \cline{4-7} 
                       &                       &                       & PointNet++  \cite{qi2017pointnet++}            & 35.64         & \multicolumn{1}{c|}{95.26 }         & 99.44          \\ \cline{4-7} 
                       &                       &                       & PointNet \cite{qi2017pointnet}              & 38.66         & \multicolumn{1}{c|}{90.52}         & 98.32          \\ \cline{4-7} 
                        &                       &                       & VoxNet \cite{maturana2015voxnet}              & 10.49         & \multicolumn{1}{c|}{12.53}         & 54.59          \\ \cline{4-7} 
                       &                       &                       & \textbf{Proposed   Method} & \textbf{5.75} & \multicolumn{1}{c|}{\textbf{2.22}} & \textbf{7.24}  \\ \cline{2-7} 
                       & \multirow{4}{*}{Wrap} & \multirow{4}{*}{Wrap} & Point-MAE \cite{pang2022MAE}                  & 9.46          & \multicolumn{1}{c|}{8.91}          & 15.87          \\ \cline{4-7} 
                       &                       &                       & PointNet++  \cite{qi2017pointnet++}            & 47.93         & \multicolumn{1}{c|}{91.36}         & 96.14          \\ \cline{4-7} 
                       &                       &                       & PointNet   \cite{qi2017pointnet}             & 5.52          & \multicolumn{1}{c|}{4.17}          & 6.12            \\ \cline{4-7} 
                        &                       &                       & VoxNet \cite{maturana2015voxnet}             &    4.18        & \multicolumn{1}{c|}{1.67}          & 4.17           \\ \cline{4-7} 
                       &                       &                       & \textbf{Proposed   Method} & \textbf{0.25} & \multicolumn{1}{c|}{\textbf{0}}    & \textbf{0.27}  \\ \hline
\multirow{8}{*}{Inter} & \multirow{4}{*}{Mask} & \multirow{4}{*}{Wrap} & Point-MAE  \cite{pang2022MAE}                 & 16.18         & \multicolumn{1}{c|}{23.11}         & 42.16          \\ \cline{4-7} 
                       &                       &                       & PointNet++  \cite{qi2017pointnet++}            & 27.24         & \multicolumn{1}{c|}{69.35}         & 80.50          \\ \cline{4-7} 
                       &                       &                       & PointNet \cite{qi2017pointnet}              & 35.87         & \multicolumn{1}{c|}{88.32}         & 92.24          \\ \cline{4-7} 
                       &                       &                       & VoxNet \cite{maturana2015voxnet}              & 14.73         & \multicolumn{1}{c|}{20.89}         & 32.13          \\ \cline{4-7} 
                       &                       &                       & \textbf{Proposed   Method} & \textbf{7.78} & \multicolumn{1}{c|}{\textbf{5.57}} & \textbf{11.14} \\ \cline{2-7} 
                       & \multirow{4}{*}{Wrap} & \multirow{4}{*}{Mask} & Point-MAE   \cite{pang2022MAE}               & 17.43         & \multicolumn{1}{c|}{25.34}         & 44.56          \\ \cline{4-7} 
                       &                       &                       & PointNet++  \cite{qi2017pointnet++}           & 14.28         & \multicolumn{1}{c|}{47.63}         & 91.36          \\ \cline{4-7} 
                       &                       &                       & PointNet  \cite{qi2017pointnet}             & 21.61         & \multicolumn{1}{c|}{30.91}         & 40.11          \\ \cline{4-7} 
                        &                       &                       & VoxNet \cite{maturana2015voxnet}             & 12.31         & \multicolumn{1}{c|}{17.27}         & 20.89          \\ \cline{4-7} 
                       &                       &                       & \textbf{Proposed   Method} & \textbf{9.24} & \multicolumn{1}{c|}{\textbf{7.79}} & \textbf{20.61} \\ \hline
\multirow{4}{*}{Both} &
  \multirow{4}{*}{Mask \& Wrap} &
  \multirow{4}{*}{Mask \&   Wrap} &
  Point-MAE \cite{pang2022MAE}  &
  11.84 &
  \multicolumn{1}{c|}{14.48} &
  22.28 \\ \cline{4-7} 
                       &                       &                       & PointNet++  \cite{qi2017pointnet++}           & 37.12         & \multicolumn{1}{c|}{90.25}         & 95.82          \\ \cline{4-7} 
                       &                       &                       & PointNet \cite{qi2017pointnet}              & 3.46          & \multicolumn{1}{c|}{1.11}          & 2.51           \\ \cline{4-7} 
                       &                       &                       & VoxNet \cite{maturana2015voxnet}                & 7.53          & \multicolumn{1}{c|}{6.96}          & 9.19          \\ \cline{4-7} 
                       &                       &                       & \textbf{Proposed   Method} & \textbf{1.42} & \multicolumn{1}{c|}{\textbf{0}}    & \textbf{0.27}  \\ \hline
\end{tabular}%
}
\end{table*}


\begin{figure*}[htp] 
    \centering
    \subfloat[PointNet \cite{qi2017pointnet} ]{%
        \includegraphics[width=0.2\textwidth]{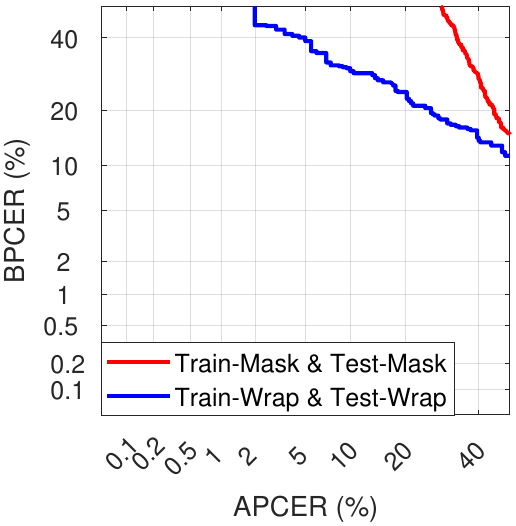}%
        \label{fig:Inter_a}%
        }%
\hfill%
      \subfloat[PointNet++ \cite{qi2017pointnet++}  ]{%
        \includegraphics[width=0.2\textwidth]{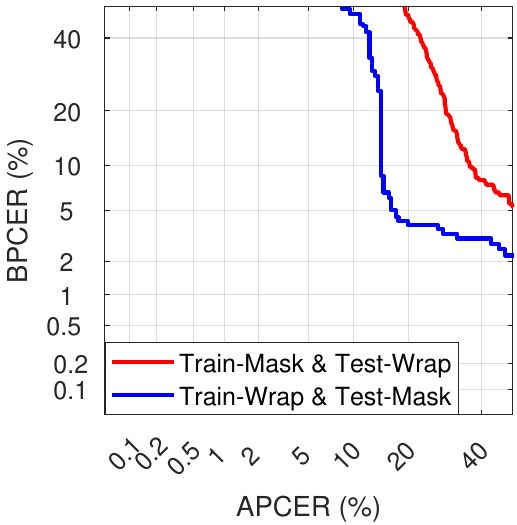}%
        \label{fig:Inter_b}%
        }%
        \hfill%
      \subfloat[Point-MAE \cite{pang2022MAE} ]{%
        \includegraphics[width=0.2\textwidth]{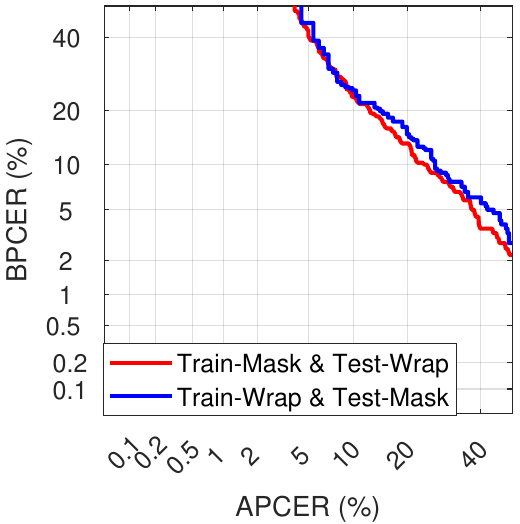}%
        \label{fig:Inter_c}%
        }%
        \hfill%
      \subfloat[VoxNet \cite{maturana2015voxnet}]{%
        \includegraphics[width=0.2\textwidth]{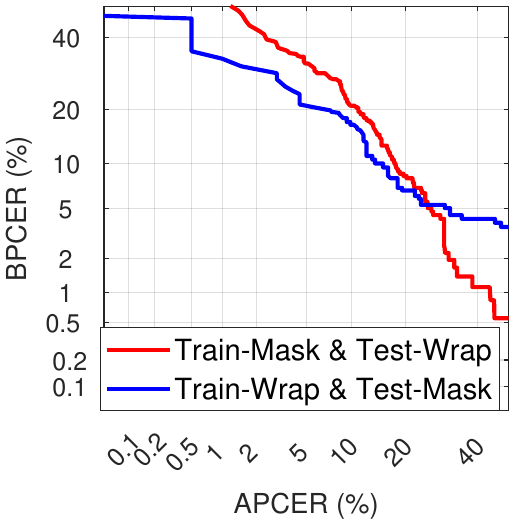}%
        \label{fig:Inter_d}%
        }%
    \hfill%
    \subfloat[Proposed Method]{%
        \includegraphics[width=0.2\textwidth]{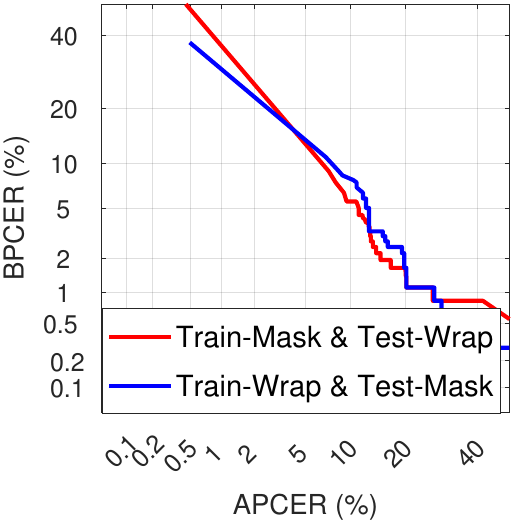}%
        \label{fig:Inter_e}%
        }%
    \caption{DET curves showing the detection performance with Inter protocol (best viewed in color). X-Axis indicates the APCER and y-axis indicates the BPCER.}
\end{figure*}

To effectively evaluate the proposed and SOTA 3D point-cloud techniques for face PAD, we propose three different evaluation protocols. The \textbf{intra protocol} trains and tests 3D point clouds with the same type of PAI. This protocol benchmarks the performance of PAD techniques with a known type of PAI. The \textbf{Inter protocol} trains the PAD techniques with one type of PAI and tests with another type of PAI. This protocol indicates the performance of PAD techniques with unseen attacks. The \textbf{Both protocol}, in which both PAIs are used for the training and testing of PADs.  This protocol benchmarks the performance of the PAD techniques when different types of PAIs are used for training and testing.

\subsection{Results and discussion}
Table \ref{tab:Resu} shows the quantitative performance of the proposed VoxAtnNet and existing methods for different performance evaluation protocols. Figure \ref{fig:Intra_a}, \ref{fig:Intra_b}, \ref{fig:Intra_c}, \ref{fig:Intra_d} and \ref{fig:Intra_e}  show the DET curves of  the proposed  and existing method  for the intra-protocol.  The following are the important observations of the intra-protocol results: 
\begin{itemize}

\item Among the two different PAI, the detection accuracy of the 3D silicone face artefact is degraded compared to 3D wrap photo Artefacts. Degraded performance can be noted with both PointNet \cite{qi2017pointnet} and the proposed method. The challenge in detecting a 3D silicone mask can be attributed to the similar facial structure and geometry of the bona fide face. However, the use of 3D wrap photos can   only generate the pseudo depth, as it is used to cover a photo that does not reflect the actual depth and shape of the actual face. This can be observed in Figure \ref{fig:Vox} in which the wrap attacks can only simulate the pseudo depth and do not provide the depth related to facial features (e.g., nose, mouth). 
\item The proposed method shows the better performance compared to the  existing methods on both type of PAIs employed in this work. The improved performance can be attributed to the use of voxelization as a feature that can effectively encode the  spatial structure that reflects the discontinuity in facial geometry in PAIs. 
\item The proposed method has indicated the D-EER = 5.75\% on 3D silicone mask and D-EER = 0.25\% on 3D wrap photo PAI on intra protocol. 
\item Among the existing methods, the VoxNet \cite{maturana2015voxnet} indicates the better results compared to the PointNet variants.
 
\end{itemize}

\begin{figure}[htp]
\begin{center}
\includegraphics[width=0.8\linewidth]{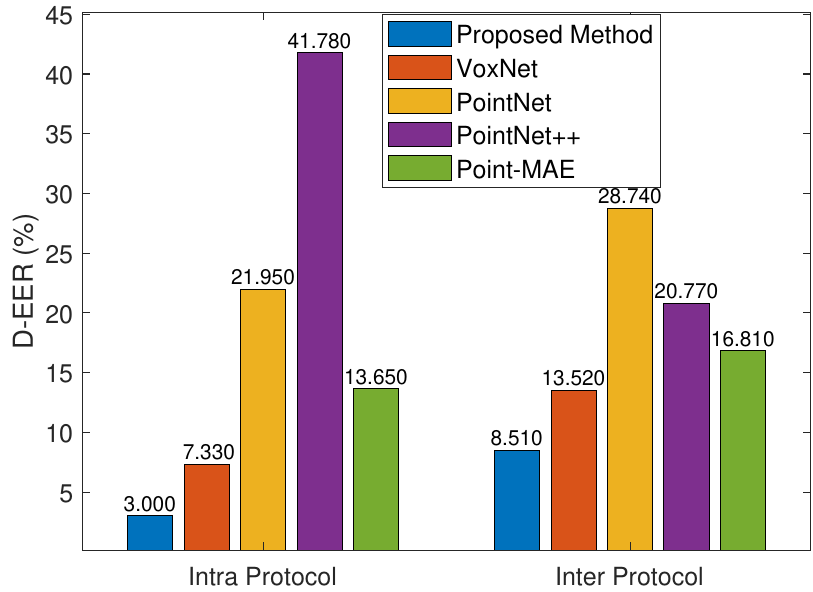}
\end{center}
   \caption{Average D-EER(\%) of the proposed method and existing methods for inter and intra protocol. The average D-EER(\%) is computed by taking the average of D-EER of the proposed or existing methods independently within the intra and inter protocol. }
\label{fig:Bar}
\end{figure}

\begin{figure}[htp]
\begin{center}
\includegraphics[width=0.8\linewidth]{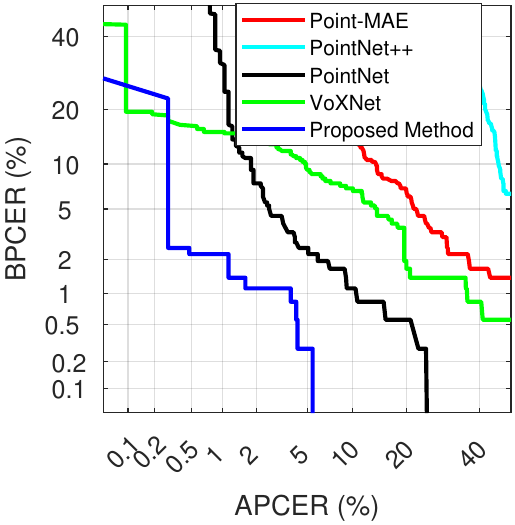}
\end{center}
   \caption{DET curves showing the detection performance with Both protocol (best viewed in color)}
\label{fig:Both}
\end{figure}

Figure \ref{fig:Inter_a}, \ref{fig:Inter_b}, \ref{fig:Inter_c}, \ref{fig:Inter_d} and \ref{fig:Inter_e} show the DET curves indicating the detection performance of  the proposed and existing methods on the inter-protocol. Based on the obtained results, as listed in Table \ref{tab:Resu}, the following can be noted:

\begin{itemize}
\item The PointNet \cite{qi2017pointnet} technique indicates the degraded performance compared to the proposed method. Training with a 3D silicone mask and testing with 3D wrap photo PAI indicates a higher degradation that can be attributed to the different types of 3D information between PAIs. 
\item The proposed method also indicates a slight drop in the detection performance in the inter evaluation protocol. As shown in the Figure \ref{fig:Bar} that average D-EER variation of the proposed method between intra and inter  evaluation protocol is less compared to the existing methods.  Thus, the proposed method has higher generalizability than existing methods. 
\end{itemize}

Figure \ref{fig:Both} shows the DET curves for `Both protocol', where both PAIs were used for training and testing. It is interesting to note that the performance of PointNet \cite{qi2017pointnet} shows a larger improvement in detection accuracy. This improved performance is due to the large number of 3D wrap PAI (857) compared to the 3D mask PAI (202). This is because PointNet \cite{qi2017pointnet} has shown good accuracy in detecting 3D wrap PAI over 3D mask PAI, as discussed above and shown in Table \ref{tab:Resu}. The proposed method exhibited the best performance for both protocols, with D-EER = 1.42\%. 

\begin{figure}[htp]
\begin{center}
\includegraphics[width=1\linewidth]{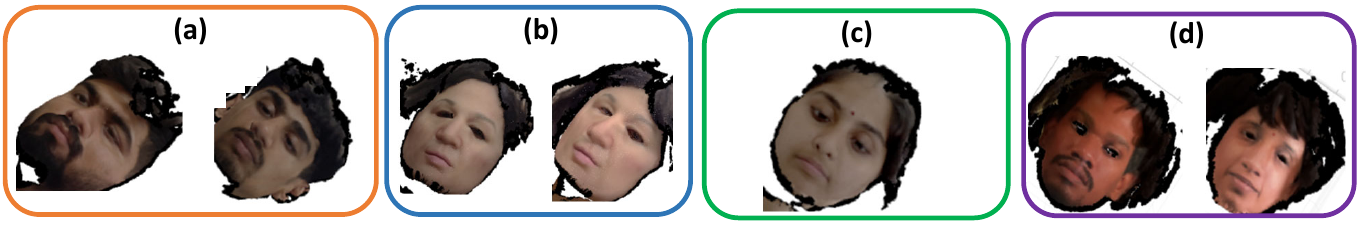}
\end{center}
   \caption{Misclassification results of the proposed method: (a) Bona fide classified as Silicone Mask (b) Silicone mask classified as Bona fide (c) Bona fide classified as Wrap Attack (d) Wrap Attack classified as Bona fide}
\label{fig:Mis}
\end{figure}

Based on the series of experiments reported above, the proposed method demonstrated the best performance across all three performance evaluation protocols. The proposed method also exhibits a smaller drop in performance, indicating the generalizable characteristics of the proposed method compared to existing methods. 

 Misclassification examples from the proposed method, as shown in Figure \ref{fig:Mis}, and evaluated using the intra-protocol, are illustrated. It is important to note that Figure \ref{fig:Mis} (c) depicts only one example because the proposed method leads to a single misclassification. Although it is difficult to draw definitive conclusions, we conclude that the environmental noise present during the self-scanning of the face might have contributed to the generation of low-quality point clouds, ultimately leading to misclassification.
\subsection{Ablation Study}
 In this section, we present an ablation study that examines the impact of various parameters and the roles of the attention module in the proposed VoxAtnNet. Table \ref{tab:Ab1} presents a hyper-parameter study that investigates different sizes of convolution filters within the VoxAtnNet framework. Three distinct filter sizes, namely $3 \times 3$, $5 \times 5$, and $7 \times 7$, were evaluated using the proposed VoxAtnNet architecture. The quantitative outcomes for the intra-evaluation protocol pertaining to mask, wrap, and both are presented in Table \ref{tab:Ab1}. The obtained results suggest that a filter size of $3 \times 3$ exhibits superior detection accuracy when confronted with wrap attacks. Conversely, a filter size of $5 \times 5$ demonstrates strong performance in detecting silicone mask attacks, and a filter size of $7 \times 7$ did not indicate good detection accuracy on detecting PA compared to other filter sizes.  The VoxAtnNet architecture was designed to incorporate both $3 \times 3$ and $5 \times 5$ filters, which demonstrates its effectiveness in detecting both silicone mask and wrap attacks, thereby validating the appropriateness of the selected filter size.

 Table \ref{tab:Ab2} presents an ablation study of the proposed VoxAtnNet, comparing its performance with and without the attention module. The results indicated that the inclusion of the attention module significantly improved the detection accuracy and generalizability of the proposed model. Without the attention module, the network's performance is severely degraded, particularly in wrap attacks, resulting in highly reduced detection accuracy and generalizability to unseen 3D PAIs.

\begin{table}[htp]
\centering
\caption{Ablation study with different parameters of convolution size in the proposed VoxAtnNet.}
\label{tab:Ab1}
\resizebox{\columnwidth}{!}{%
\begin{tabular}{|c|c|c|ccc|}
\hline
\multirow{3}{*}{\textbf{Ablation Parameters}} &
  \multirow{3}{*}{\textbf{Train}} &
  \multirow{3}{*}{\textbf{Test}} &
  \multicolumn{3}{c|}{\textbf{Detection Performance}} \\ \cline{4-6} 
                                      &      &      & \multicolumn{1}{c|}{\multirow{2}{*}{\textbf{D-EER (\%)}}} & \multicolumn{2}{c|}{\textbf{BPCER at APCER =}}     \\ \cline{5-6} 
                                      &      &      & \multicolumn{1}{c|}{}                                     & \multicolumn{1}{c|}{\textbf{10\%}} & \textbf{5\%}  \\ \hline
\multirow{3}{*}{Convolution with 3X3} & Mask & Mask & \multicolumn{1}{c|}{29.75}                                & \multicolumn{1}{c|}{74.93}         & 83.28         \\ \cline{2-6} 
                                      & Wrap & Wrap & \multicolumn{1}{c|}{1.39}                                 & \multicolumn{1}{c|}{0.55}          & 1.39          \\ \cline{2-6} 
                                      & Both & Both & \multicolumn{1}{c|}{3.68}                                 & \multicolumn{1}{c|}{2.78}          & 3.16          \\ \hline
\multirow{3}{*}{Convolution with 5X5} & Mask & Mask & \multicolumn{1}{c|}{9.43}                                 & \multicolumn{1}{c|}{8.91}          & 45.96         \\ \cline{2-6} 
                                      & Wrap & Wrap & \multicolumn{1}{c|}{4.44}                                 & \multicolumn{1}{c|}{1.94}          & 4.17          \\ \cline{2-6} 
                                      & Both & Both & \multicolumn{1}{c|}{2.87}                                 & \multicolumn{1}{c|}{1.11}          & 1.11          \\ \hline
\multirow{3}{*}{Convolution with 7X7} & Mask & Mask & \multicolumn{1}{c|}{12.84}                                & \multicolumn{1}{c|}{15.32}         & 19.49         \\ \cline{2-6} 
                                      & Wrap & Wrap & \multicolumn{1}{c|}{18.13}                                & \multicolumn{1}{c|}{48.18}         & 64.92         \\ \cline{2-6} 
                                      & Both & Both & \multicolumn{1}{c|}{4.73}                                 & \multicolumn{1}{c|}{3.89}          & 4.73          \\ \hline
\multirow{3}{*}{\textbf{\begin{tabular}[c]{@{}c@{}}Proposed Method: \\ Combining 3X3 and 5X5\end{tabular}}} &
  Mask &
  Mask &
  \multicolumn{1}{c|}{\textbf{5.75}} &
  \multicolumn{1}{c|}{\textbf{2.22}} &
  \textbf{7.24} \\ \cline{2-6} 
                                      & Wrap & Wrap & \multicolumn{1}{c|}{\textbf{0.25}}                        & \multicolumn{1}{c|}{\textbf{0}}    & \textbf{0}    \\ \cline{2-6} 
                                      & Both & Both & \multicolumn{1}{c|}{\textbf{1.42}}                        & \multicolumn{1}{c|}{\textbf{0}}    & \textbf{0.27} \\ \hline
\end{tabular}%
}
\end{table}

\begin{table}[]
\centering
\caption{Ablation study with/without proposed attention}
\label{tab:Ab2}
\resizebox{1\columnwidth}{!}{%
\begin{tabular}{|c|c|c|ccc|}
\hline
\multirow{3}{*}{\textbf{Ablation Parameters}} &
  \multirow{3}{*}{\textbf{Train}} &
  \multirow{3}{*}{\textbf{Test}} &
  \multicolumn{3}{c|}{\textbf{Detection Performance}} \\ \cline{4-6} 
                                   &      &      & \multicolumn{1}{c|}{\multirow{2}{*}{\textbf{D-EER (\%)}}} & \multicolumn{2}{c|}{\textbf{BPCER ' APCER =}}      \\ \cline{5-6} 
                                   &      &      & \multicolumn{1}{c|}{}                                     & \multicolumn{1}{c|}{\textbf{10\%}} & \textbf{5\%}  \\ \hline
\multirow{3}{*}{Without Attention} & Mask & Mask & \multicolumn{1}{c|}{20.84}                                & \multicolumn{1}{c|}{33.14}         & 38.44         \\ \cline{2-6} 
                                   & Wrap & Wrap & \multicolumn{1}{c|}{50}                                   & \multicolumn{1}{c|}{80.50}         & 86.62         \\ \cline{2-6} 
                                   & Both & Both & \multicolumn{1}{c|}{12.31}                                & \multicolumn{1}{c|}{16.71}         & 29.24         \\ \hline
\multirow{3}{*}{\textbf{\begin{tabular}[c]{@{}c@{}}With Attention\\  (Proposed Method)\end{tabular}}} &
  Mask &
  Mask &
  \multicolumn{1}{c|}{\textbf{5.75}} &
  \multicolumn{1}{c|}{\textbf{2.22}} &
  \textbf{7.24} \\ \cline{2-6} 
                                   & Wrap & Wrap & \multicolumn{1}{c|}{\textbf{0.25}}                        & \multicolumn{1}{c|}{\textbf{0}}    & \textbf{0}    \\ \cline{2-6} 
                                   & Both & Both & \multicolumn{1}{c|}{\textbf{1.42}}                        & \multicolumn{1}{c|}{\textbf{0}}    & \textbf{0.27} \\ \hline
\end{tabular}%
}
\end{table}

\section{Limitations}
\label{sec:limit}
In this work, we introduced a novel method that leverages 3D point clouds from the frontal camera of the Apple iPhone 12 Pro to detect face PAs. We demonstrate the efficacy of the proposed VoxAtnNet through a series of evaluation protocols for 3D PAIs. However, it should be noted that we collected a dataset with 3480 3D point clouds in an indoor office environment. Therefore, it is essential to benchmark the performance on a large-scale dataset by considering other lighting conditions that can reflect real-life scenarios. Furthermore, the dataset must be scaled with other types of smartphones with depth sensor (apart from iPhone) to benchmark the robustness of the proposed VoxAtnNet better. The experimental results show the pseudo depth introduced by 3D wrap photo PAI, which is less challenging to detect using the proposed VoxAtnNet. Therefore, the generalizability of PAD techniques must be extended to other 3D face masks that can reflect facial geometry. 
\section{Conclusion}
\label{sec:Cocn}
In this work, we presented a novel 3D face PAD algorithm for detecting presentation attacks on smartphones. An integral part of the proposed method is voxelization to capture the spatial structure from the point clouds, and the novel attention framework presented together with serial convolution layers.  The proposed VoxAtnNet processes the input 3D point clouds from the iPhone, followed by Voxelization,  and passing them through 3D convolutional network with 23 layers and an overall of  35.7 million learnable parameters.  In this work, we also introduce a new 3D face point cloud presentation attack dataset (3D-PCPA) with 3480 samples corresponding to Bona fide, 3D silicone mask PAI, and 3D wrap photo PAI. Extensive experiments were presented with three different performance evaluation protocols that indicate improved attack detection performance compared with four different existing methods. The obtained results indicate the superior performance of the proposed method, thus indicating a generalized PAD using 3D point clouds captured from the smartphone.

{\small
\bibliographystyle{ieee}
\bibliography{egbib}

\begin{thebibliography}{10}\itemsep=-1pt

\bibitem{SmarptMar}
{SmartPhone Market}.
\newblock
  \url{https://www.biometricupdate.com/202001/biometric-facial-recognition-hardware-present-in-90-of-smartphones-by-2024},
  2020.

\bibitem{SmrtSpoof}
{SmartPhone Spoofing}.
\newblock
  \url{https://press.which.co.uk/whichpressreleases/smartphones-have-face-recognition-that-can-be-easily-spoofed-with-2d-photo-which-finds/},
  2020.

\bibitem{atoum2017face}
Y.~Atoum, Y.~Liu, A.~Jourabloo, and X.~Liu.
\newblock Face anti-spoofing using patch and depth-based cnns.
\newblock In {\em IJCB}, pages 319--328. IEEE, 2017.

\bibitem{boulkenafet2016face}
Z.~Boulkenafet, J.~Komulainen, and A.~Hadid.
\newblock Face antispoofing using speeded-up robust features and fisher vector
  encoding.
\newblock {\em IEEE Signal Processing Letters}, 24(2):141--145, 2016.

\bibitem{boulkenafet2016face1}
Z.~Boulkenafet, J.~Komulainen, and A.~Hadid.
\newblock Face spoofing detection using colour texture analysis.
\newblock {\em TIFS}, 11(8):1818--1830, 2016.

\bibitem{Moier2015face}
D.~C. Garcia and R.~L. de~Queiroz.
\newblock Face-spoofing 2d-detection based on moir{\'e}-pattern analysis.
\newblock {\em IEEE transactions on information forensics and security},
  10(4):778--786, 2015.

\bibitem{george2019deep}
A.~George and S.~Marcel.
\newblock Deep pixel-wise binary supervision for face presentation attack
  detection.
\newblock In {\em ICB}, pages 1--8. IEEE, 2019.

\bibitem{george2019biometric}
A.~George, Z.~Mostaani, D.~Geissenbuhler, O.~Nikisins, A.~Anjos, and S.~Marcel.
\newblock Biometric face presentation attack detection with multi-channel
  convolutional neural network.
\newblock {\em IEEE Transactions on Information Forensics and Security},
  15:42--55, 2019.

\bibitem{TransformerPAD}
P.-K. Huang, C.-H. Chiang, J.-X. Chong, T.-H. Chen, H.-Y. Ni, and C.-T. Hsu.
\newblock Ldcformer: Incorporating learnable descriptive convolution to vision
  transformer for face anti-spoofing.
\newblock In {\em 2023 IEEE International Conference on Image Processing
  (ICIP)}, pages 121--125, 2023.

\bibitem{ISO-IEC-30107-3-PAD-metrics-170227}
{ISO/IEC JTC1 SC37 Biometrics}.
\newblock {\em {ISO/IEC} 30107-3. Information Technology - Biometric
  presentation attack detection - Part 3: Testing and Reporting}.
\newblock International Organization for Standardization, 2017.

\bibitem{PointcloudnetPAD}
Y.~Kim, H.~Gwak, J.~Oh, M.~Kang, J.~Kim, H.~Kwon, and S.~Kim.
\newblock Cloudnet: A lidar-based face anti-spoofing model that is robust
  against light variation.
\newblock {\em IEEE Access}, 11:16984--16993, 2023.

\bibitem{SelfSuperPADRaghu}
Z.~Kong, W.~Zhang, F.~Liu, W.~Luo, H.~Liu, L.~Shen, and R.~Ramachandra.
\newblock Taming self-supervised learning for presentation attack detection:
  De-folding and de-mixing.
\newblock {\em IEEE Transactions on Neural Networks and Learning Systems},
  pages 1--12, 2023.

\bibitem{li20203dpc}
X.~Li, J.~Wan, Y.~Jin, A.~Liu, G.~Guo, and S.~Z. Li.
\newblock 3dpc-net: 3d point cloud network for face anti-spoofing.
\newblock In {\em 2020 IEEE International Joint Conference on Biometrics
  (IJCB)}, pages 1--8. IEEE, 2020.

\bibitem{liu2022contrastive}
A.~Liu, C.~Zhao, Z.~Yu, J.~Wan, A.~Su, X.~Liu, Z.~Tan, S.~Escalera, J.~Xing,
  Y.~Liang, et~al.
\newblock Contrastive context-aware learning for 3d high-fidelity mask face
  presentation attack detection.
\newblock {\em IEEE Transactions on Information Forensics and Security},
  17:2497--2507, 2022.

\bibitem{liu2018learning}
Y.~Liu, A.~Jourabloo, and X.~Liu.
\newblock Learning deep models for face anti-spoofing: Binary or auxiliary
  supervision.
\newblock In {\em CVPR}, pages 389--398, 2018.

\bibitem{maatta2011face}
J.~M{\"a}{\"a}tt{\"a}, A.~Hadid, and M.~Pietik{\"a}inen.
\newblock Face spoofing detection from single images using micro-texture
  analysis.
\newblock In {\em IJCB}, pages 1--7. IEEE, 2011.

\bibitem{HOGPAD}
J.~M{\"a}{\"a}tt{\"a}, A.~Hadid, and M.~Pietik{\"a}inen.
\newblock Face spoofing detection from single images using texture and local
  shape analysis.
\newblock {\em IET biometrics}, 1(1):3--10, 2012.

\bibitem{maturana20153d}
D.~Maturana and S.~Scherer.
\newblock 3d convolutional neural networks for landing zone detection from
  lidar.
\newblock In {\em 2015 IEEE international conference on robotics and automation
  (ICRA)}, pages 3471--3478. IEEE, 2015.

\bibitem{maturana2015voxnet}
D.~Maturana and S.~Scherer.
\newblock Voxnet: A 3d convolutional neural network for real-time object
  recognition.
\newblock In {\em 2015 IEEE/RSJ international conference on intelligent robots
  and systems (IROS)}, pages 922--928. IEEE, 2015.

\bibitem{pang2022MAE}
Y.~Pang, W.~Wang, F.~E. Tay, W.~Liu, Y.~Tian, and L.~Yuan.
\newblock Masked autoencoders for point cloud self-supervised learning.
\newblock In {\em European conference on computer vision}, pages 604--621.
  Springer, 2022.

\bibitem{patel2016secure}
K.~Patel, H.~Han, and A.~K. Jain.
\newblock Secure face unlock: Spoof detection on smartphones.
\newblock {\em TIFS}, 11(10):2268--2283, 2016.

\bibitem{qi2017pointnet}
C.~R. Qi, H.~Su, K.~Mo, and L.~J. Guibas.
\newblock Pointnet: Deep learning on point sets for 3d classification and
  segmentation.
\newblock In {\em Proceedings of the IEEE conference on computer vision and
  pattern recognition}, pages 652--660, 2017.

\bibitem{qi2017pointnet++}
C.~R. Qi, L.~Yi, H.~Su, and L.~J. Guibas.
\newblock Pointnet++: Deep hierarchical feature learning on point sets in a
  metric space.
\newblock {\em Advances in neural information processing systems}, 30, 2017.

\bibitem{BSIFFacePAD}
R.~Raghavendra and C.~Busch.
\newblock Presentation attack detection algorithm for face and iris biometrics.
\newblock In {\em 2014 22nd European signal processing conference (EUSIPCO)},
  pages 1387--1391. IEEE, 2014.

\bibitem{ramachandra2017presentation}
R.~Ramachandra and C.~Busch.
\newblock Presentation attack detection methods for face recognition systems: A
  comprehensive survey.
\newblock {\em CSUR}, 50(1):1--37, 2017.

\bibitem{ramachandra2020face}
R.~Ramachandra, J.~M. Singh, S.~Venkatesh, K.~Raja, and C.~Busch.
\newblock Face presentation attack detection using multi-classifier fusion of
  off-the-shelf deep features.
\newblock In {\em Computer Vision and Image Processing: 4th International
  Conference, CVIP 2019, Jaipur, India, September 27--29, 2019, Revised
  Selected Papers, Part II 4}, pages 49--61. Springer, 2020.

\bibitem{shao2019multi}
R.~Shao, X.~Lan, J.~Li, and P.~C. Yuen.
\newblock Multi-adversarial discriminative deep domain generalization for face
  presentation attack detection.
\newblock In {\em CVPR}, pages 10023--10031, 2019.

\bibitem{shao2020regularized}
R.~Shao, X.~Lan, and P.~C. Yuen.
\newblock Regularized fine-grained meta face anti-spoofing.
\newblock In {\em Proceedings of the AAAI Conference on Artificial
  Intelligence}, volume~34, pages 11974--11981, 2020.

\bibitem{sun2022danet}
C.-Y. Sun, S.-L. Chen, X.-J. Li, F.~Chen, and X.-C. Yin.
\newblock Danet: Dynamic attention to spoof patterns for face anti-spoofing.
\newblock In {\em 2022 26th International Conference on Pattern Recognition
  (ICPR)}, pages 1929--1936. IEEE, 2022.

\bibitem{wang2020cross}
G.~Wang, H.~Han, S.~Shan, and X.~Chen.
\newblock Cross-domain face presentation attack detection via multi-domain
  disentangled representation learning.
\newblock In {\em CVPR}, pages 6678--6687, 2020.

\bibitem{wang2019multi}
G.~Wang, C.~Lan, H.~Han, S.~Shan, and X.~Chen.
\newblock Multi-modal face presentation attack detection via spatial and
  channel attentions.
\newblock In {\em Proceedings of the IEEE/CVF Conference on Computer Vision and
  Pattern Recognition Workshops}, pages 0--0, 2019.

\bibitem{wang2020deep}
Z.~Wang, Z.~Yu, C.~Zhao, X.~Zhu, Y.~Qin, Q.~Zhou, F.~Zhou, and Z.~Lei.
\newblock Deep spatial gradient and temporal depth learning for face
  anti-spoofing.
\newblock In {\em Proceedings of the IEEE/CVF Conference on Computer Vision and
  Pattern Recognition}, pages 5042--5051, 2020.

\bibitem{yang2014learn}
J.~Yang, Z.~Lei, and S.~Z. Li.
\newblock Learn convolutional neural network for face anti-spoofing.
\newblock {\em arXiv preprint arXiv:1408.5601}, 2014.

\bibitem{yu2021salience}
B.~Yu, J.~Lu, X.~Li, and J.~Zhou.
\newblock Salience-aware face presentation attack detection via deep
  reinforcement learning.
\newblock {\em IEEE Transactions on Information Forensics and Security},
  17:413--427, 2021.

\bibitem{yu2021transrppg}
Z.~Yu, X.~Li, P.~Wang, and G.~Zhao.
\newblock Transrppg: Remote photoplethysmography transformer for 3d mask face
  presentation attack detection.
\newblock {\em IEEE Signal Processing Letters}, 28:1290--1294, 2021.

\bibitem{yu2022deep}
Z.~Yu, Y.~Qin, X.~Li, C.~Zhao, Z.~Lei, and G.~Zhao.
\newblock Deep learning for face anti-spoofing: A survey.
\newblock {\em IEEE Transactions on Pattern Analysis and Machine Intelligence
  (TPAMI)}, 2022.

\bibitem{zhang2021face}
W.~Zhang, H.~Liu, F.~Liu, R.~Ramachandra, and C.~Busch.
\newblock Effective presentation attack detection driven by face related task.
\newblock In {\em 17th European Conference of Computer Vision (ECCV)}, pages
  408--423. Springer, 2022.

\bibitem{zhang2022effective}
W.~Zhang, H.~Liu, F.~Liu, R.~Ramachandra, and C.~Busch.
\newblock Effective presentation attack detection driven by face related task.
\newblock In {\em European Conference on Computer Vision}, pages 408--423.
  Springer, 2022.

\end{thebibliography}
}

\end{document}